\documentclass[runningheads]{llncs}
\usepackage[T1]{fontenc}
\usepackage{amsmath,graphicx,lipsum}

\usepackage{listings,xfp}
\usepackage{booktabs}
\usepackage{xcolor}
\usepackage{enumitem}
\usepackage{tikz}
\usetikzlibrary{shapes.geometric}
\usepackage{wrapfig}
\usepackage{pgfplots}
\usepackage{float}
\usetikzlibrary{positioning}
\usepackage{subcaption}
\usepackage{tikz,pgfplots}
\usepackage{epsfig}
\usetikzlibrary{pgfplots.groupplots}
\usetikzlibrary{patterns}

\usepackage{graphicx}
\usepackage{amsmath}
\usepackage{amssymb}
\usepackage{booktabs}
\usepackage{cite}
\usepackage{amsmath,amssymb,amsfonts}
\usepackage{algorithmic}
\usepackage{booktabs}
\usepackage[pagebackref,breaklinks,colorlinks]{hyperref}
\usepackage{graphicx}
\usepackage{tikz}
\usepackage{textcomp}
\usepackage{xcolor}
\def\BibTeX{{\rm B\kern-.05em{\sc i\kern-.025em b}\kern-.08em
    T\kern-.1667em\lower.7ex\hbox{E}\kern-.125emX}}
\usepackage{graphicx}
\begin{document}
\title{Object-Centric Cropping for
Visual Few-Shot Classification}
%
%
\author{Aymane Abdali\inst{1,2,*} \and
Bartosz Boguslawski\inst{2} \and
Lucas Drumetz\inst{1} \and
Vincent Gripon\inst{1}}
%

\institute{IMT Atlantique, UMR CNRS 6285, Lab-STICC, F-29238 Brest, France 
\and
Schneider Electric, Grenoble, France\\
\email{\{*\} aymane.abdali@gmail.com}}
\maketitle              

\begin{abstract}
In the domain of Few-Shot Image Classification, operating with as little as one example per class, the presence of image ambiguities stemming from multiple objects or complex backgrounds can significantly deteriorate performance. 
Our research demonstrates that incorporating additional information about the local positioning of an object within its image markedly enhances classification across established benchmarks. 
More importantly, we show that a significant fraction of the improvement can be achieved through the use of the Segment Anything Model~\cite{kirillov2023segment}, requiring only a pixel of the object of interest to be pointed out, or by employing fully unsupervised foreground object extraction methods.

\end{abstract}
\section{Introduction}
\label{sec:intro}

Recent advancements in Few-Shot adaptation, whether through model fine-tuning or more sophisticated methods, have empowered models to rapidly learn from small datasets~\cite{bendou2022easy,chen2019closer}. Several large models pretrained on diverse image datasets, such as CLIP or DINO~\cite{radford2021learning,simeoni2021localizing}, are now accessible and deliver competitive results in classification benchmarks. Despite these models possessing intriguing adaptation and transfer capabilities, they may encounter challenges in correctly generalizing to datasets with task ambiguity~\cite{tamkin2022active}.

For instance, a model trained to classify waterbirds and landbirds may factor in the background when making decisions, potentially leading to misclassification of birds not in their typical habitats. Hence, for optimal task adaptation, it is crucial to ensure that the models learn to classify images based on relevant features.

\begin{figure*}

\scalebox{0.8}{
\begin{subfigure}[b]{1\textwidth}
\centering
\begin{tikzpicture}[thick,scale=0.9, every node/.style={scale=0.9}]

    \node(humB)[label={\fontsize{7}{10}\selectfont human annotator}] at (0,-2){\includegraphics[width=8mm]{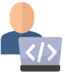}};
    \node(cropcow) at (6,-2){\includegraphics[width=2.5cm]{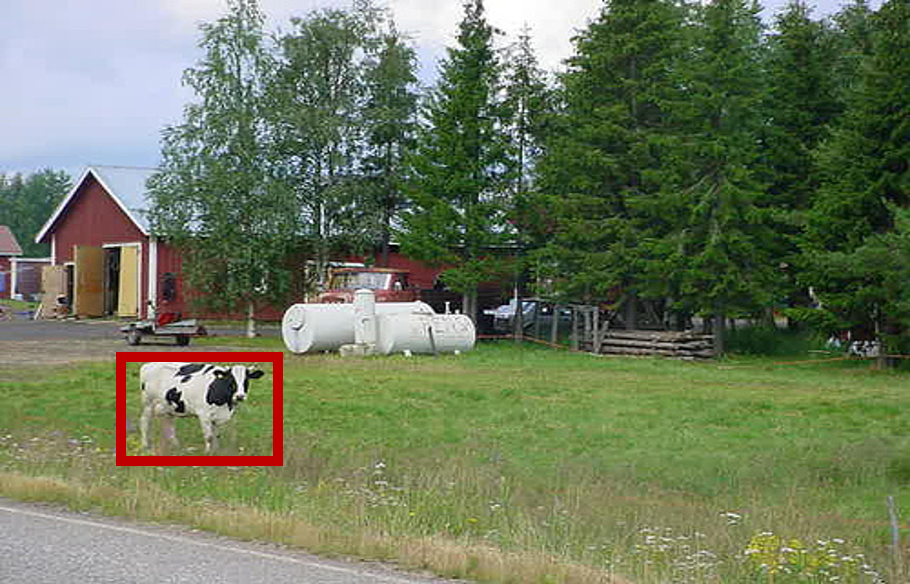}};
    \node(auggt)[label=below:{\fontsize{7}{10}\selectfont single object centric augment}] at (11,-2){\includegraphics[width=2cm]{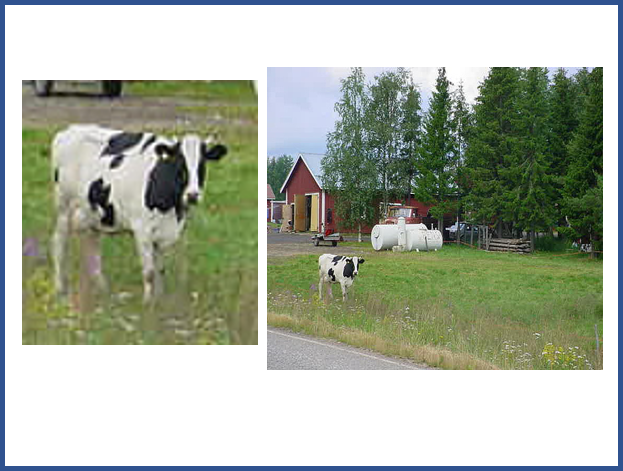}};

    \node(imenc2)[label={[rotate=-90]center:{\fontsize{7}{10}\selectfont Image Encoder}},draw, trapezium,minimum height=1cm,rotate=-90,minimum width = 3 cm,trapezium right angle=60,trapezium left angle=60]  at (14.5,-2){};]
    \node(lin2)[label={[rotate=-90]center:{\fontsize{7}{10}\selectfont Classification Head }},draw, rectangle,minimum height=0.5cm,rotate=-90,minimum width = 2.8cm]  at (15.7,-2){};]

    \draw[->] (imenc2) -- (lin2);
    \draw[->] (humB) -- (cropcow)node [below,pos=0.5] {{\fontsize{7}{10}\selectfont ``bound the object with a box''}};
    \draw[->] (cropcow) -- (auggt);
    \draw[->] (auggt) -- (imenc2);

\end{tikzpicture}
\caption{Ground truth bounding box disambiguation.}
\end{subfigure}
}
\scalebox{0.8}{
\begin{subfigure}[a]{1\textwidth}
\centering
\begin{tikzpicture}[thick,scale=1, every node/.style={scale=1}]
    \node(humA)[label={\fontsize{7}{10}\selectfont human annotator}] at (0,2){\includegraphics[width=8mm]{humann.png}};
    \node(clickow) at (3,3){\includegraphics[width=2cm]{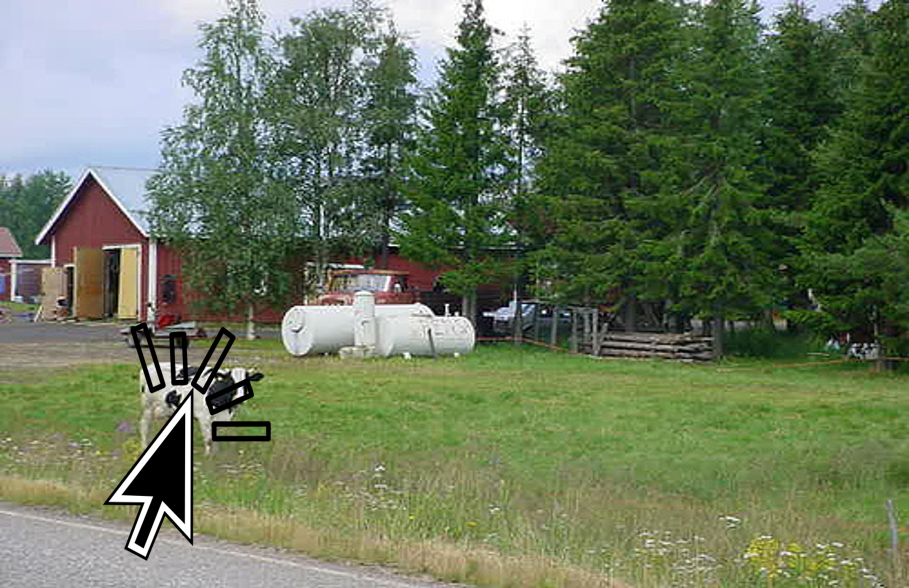}};
    \node(SAM)[draw, trapezium,minimum height=0.8cm,minimum width = 0.5cm,trapezium right angle=120]  at (5.2,2){{\fontsize{7}{10}\selectfont SAM}};
    \node(maskcow)[label={\fontsize{8}{10}\selectfont SAM}] at (8,2){\includegraphics[width=2.5cm]{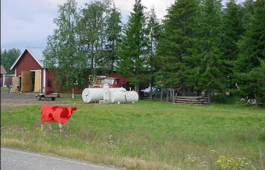}};
    \node(augsam)[label={\fontsize{7}{10}\selectfont multiple augments}] at (11,2){\includegraphics[width=2cm]{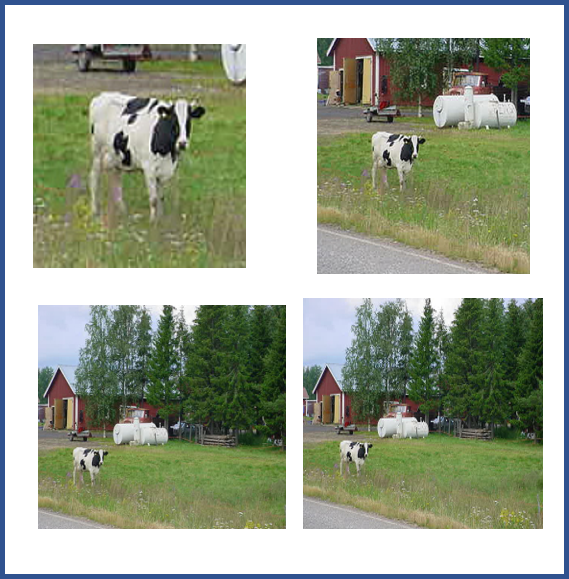}};
    \node(imenc)[label={[rotate=-90]center:{\fontsize{7}{10}\selectfont Image Encoder}},draw, trapezium,minimum height=1cm,rotate=-90,minimum width = 3 cm,trapezium right angle=60,trapezium left angle=60]  at (13,2){};]
    \node(lin)[label={[rotate=-90]center:{\fontsize{7}{10}\selectfont Classification Head }},draw, rectangle,minimum height=0.5cm,rotate=-90,minimum width = 2.8cm]  at (14.2,2){};]

    \draw[->] (humA) -- (SAM) node [below,pos=0.4] {{\fontsize{7}{10}\selectfont ``point click on object''}};
    \draw[->] (SAM) -- (maskcow);
    \draw[->] (maskcow) -- (augsam);
    \draw[->] (augsam) -- (imenc);
    \draw[->] (imenc) -- (lin);
\end{tikzpicture}
\caption{SAM-based point click disambiguation.}
\end{subfigure}
}
\scalebox{0.8}{
\begin{subfigure}[a]{1\textwidth}
\centering
\begin{tikzpicture}[thick,scale=0.9, every node/.style={scale=0.9}]
    \node(humA)[label={\fontsize{8}{10}\selectfont}] at (0,2){\includegraphics[width=25mm]{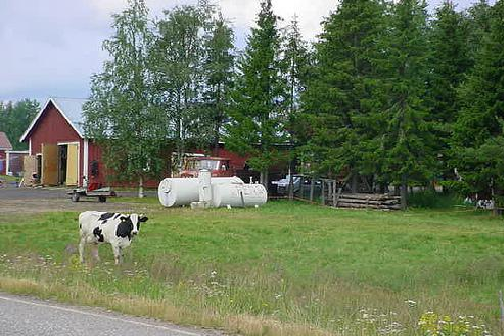}};
    \node(salcow)[label={\fontsize{8}{10}\selectfont}] at (7,2){\includegraphics[width=2.5cm]{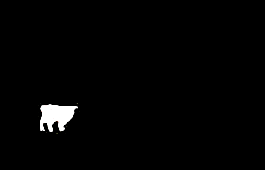}};
    \node(augsam)[label={\fontsize{7}{10}\selectfont multiple augments}] at (12,2){\includegraphics[width=2cm]{augsam.png}};
    \node(imenc)[label={[rotate=-90]center:{\fontsize{7}{10}\selectfont Image Encoder}},draw, trapezium,minimum height=1cm,rotate=-90,minimum width = 3 cm,trapezium right angle=60,trapezium left angle=60]  at (14.5,2){};]
    \node(lin)[label={[rotate=-90]center:{\fontsize{7}{10}\selectfont Classification Head }},draw, rectangle,minimum height=0.5cm,rotate=-90,minimum width = 2.8cm]  at (15.7,2){};]

    \draw[->] (humA) -- (salcow) node [below,pos=0.5] {{\fontsize{7}{10}\selectfont  ``salient object detection model''}};

    \draw[->] (salcow) -- (augsam);
    \draw[->] (augsam) -- (imenc);
    \draw[->] (imenc) -- (lin);
\end{tikzpicture}
\caption{Salient object disambiguation.}
\end{subfigure}
}
\caption{The illustration depicts the three approaches for determining the local position of the object of interest within an image. In the first scenario (a), an expert directly provides a bounding box that is then used to augment the image once. Conversely, in the second scenario (b), the expert selects a point on the object, which is subsequently employed by SAM to produce a mask. This mask is then utilized to create multiple augmentations of the image. In the third scenario (c), a mask of the relevant object is automatically generated with the help of a salient object detection model. The mask is then utilized to create multiple augmentations of the image similarly as in (b). The original image and its augmentations are processed through an image encoder then fed to a linear layer for training.}\label{illus}
\end{figure*}

Recall that in visual few-shot classification, the aim is to correctly classify objects in images using only a very limited number of examples, called \emph{shots}, for each class.
In this work, we achieve disambiguation by cropping out the exact object of interest from the shots. We consider a manual and semi-manual ways of acquiring this local position: a human annotator either provides the ground truth crop or solely indicates a pixel from the object, which is then used to prompt the Segment Anything Model (SAM)~\cite{kirillov2023segment} to automatically obtain a segmentation mask of the object. We also consider a fully unsupervised third way which consists in leveraging a salient object detection model to retrieve a mask on the object of interest. Figure~\ref{illus} depicts these three modes.
Our aim is to measure how much improvement can be achieved in classification accuracy by leveraging this local information, and to compare it with using additional shots instead. A main difficulty is to identify a suitable methodology to leverage the location of objects of interest in the shots: as a matter of fact, our research shows that focusing on the crops containing the objects of interest during learning can introduce a bias in the generalization, making it less effective. Instead, we introduce a method that integrates the crops and a portion of their context during learning, leading to consistent improvements across our experiments. Our goal is to contribute to a better understanding on how object localization can be benefit Few-Shot learning.
In this paper, we:
\begin{itemize}
  \item Propose a methodology for extracting the object of interest from a given training set to improve training.
  \item Compare multiple methods for acquiring and utilizing this information with differing precision and levels of human involvement in both inductive and transductive settings, across three datasets.
  \item Investigate the benefits of inference time disambiguation through automatic salient object detection.
  \item Examine the effects of removing the context around the object of interest of a specific class on the class distribution.
\end{itemize}

\newcommand{\rulesep}{\unskip\ \vrule height -1ex\ }

\section{Related Work}
\label{sec:rw}
\textbf{Few-Shot Learning}: The Few-Shot learning paradigm has garnered increasing attention in recent years, with notable exploration in meta-learning~\cite{nichol2018first,finn2017model,finn2018probabilistic}, aimed at acquiring task-level meta-knowledge for rapid adaptation to new tasks with minimal labeled examples. These approaches often require pre-training on more general tasks, benefiting from advancements in feature extraction, data augmentation, ensembling, and other techniques. While inductive classification (where predictions for each sample are inferred independently) has relied on simpler methods, diverse techniques have been employed in the transductive setting (inferring predictions for a pool of samples at once), such as soft K-means pseudolabeling~\cite{abdali2023active,bendou2022easy}, which we utilize in our methodology.\\
\textbf{Task Ambiguity}: The issue of task ambiguity emerges as a recurring theme in the transfer learning literature. Many studies addressing task adaptation introduce various forms of ambiguity that may arise in the new task targeted for model adaptation. Specifically, the work by \cite{finn2018probabilistic} sets out to develop a meta-learning approach capable of addressing the ambiguity inherent in learning from limited data. Furthermore, \cite{tamkin2022active} delineates task ambiguity within the context adopted in this paper, particularly in relation to extraneous objects and/or background elements coexisting within the same image. The authors of \cite{xu2024eliminating} tackle feature ambiguity by introducing the Ambiguity Elimination Network (AENet), which mines discriminative query foreground regions and rectifies ambiguous foreground features by reducing the influence of background features. This enhances foreground-to-foreground matching in cross attention and leads to improved few-shot segmentation performance. While these works provide meaningful solutions to learn despite the ambiguity, they have not considered altering the original image to mitigate it. \\
\textbf{Object Detection And Segmentation}: Segment Anything Model~\cite{kirillov2023segment,zhao2023fast,ravi2024sam} has marked a significant milestone as the inaugural large-scale segmentation model, showcasing robust zero-shot capabilities in classification and object detection tasks. Notably, this model has the capacity to generate masks in response to prompts such as points or bounding boxes. There are also notable works exploring the concept of identifying objects of interest in~\cite{hao2025simple,zhou2024admnet}. Notably, ~\cite{bielski2022move} achieves compelling results in Salient Object Detection without any supervised training. These works opens the door to a more automated approach for acquiring information about the precise location of an object of interest.

\section{Problem Formulation}
\label{sec:format}

In this work, we consider the inductive and the transductive scenarios of few-shot learning in vision tasks which are two prevalent settings in the few-shot learning literature \cite{bendou2022easy,veilleux2021realistic,hu2023adaptive}. In these two cases, we evaluate the performance gains that can be achieved through the additional information that is the local positioning of an object of interest within the image. 

\subsection{Inductive Few-Shot Classification}
\label{induc}
In inductive few-shot classification, we are given a dataset \(\mathcal{S}:= \left\{ \mathbf{x}_i^s,y_i^s \right\}^{n_s}_{i=1}\) consisting of \(n_s\) samples uniformly distributed among \(k\) classes. Here, \(\mathbf{x}_i^s\) represents the samples and \(y_i^s\) represents their respective labels. We refer to \(\mathcal{S}\) as ``the support set.'' The support set contains very few samples, typically as low as one sample per class. Additionally, we are given a test set \(\mathcal{T}:= \left\{ \mathbf{x}_i^t,y_i^t \right\}^{n_t}_{i=1}\) comprising \(n_t\) samples also uniformly distributed among the \(K\) classes. The task involves accurately predicting the labels of test samples one at time and independently from each other while having access to the label information from the support set.

\subsection{Transductive Few-Shot Classification}

In transductive few-shot classification, we are given, in addition to the support set and the test set defined in Section~\ref{induc}, a third set \(\mathcal{Q}:= \left\{ \mathbf{x}_i^q,y_i^q \right\}^{n_q}_{i=1}\) consisting of \(n_q\) samples uniformly distributed among \(K\) classes. Here, \(\mathbf{x}_i^q\) represents the samples and \(y_i^q\) represents their respective labels that are unknown. We refer to \(\mathcal{Q}\) as ``the query set.'' The query set serves as an unlabeled dataset that can provide additional information about the class distribution. This information can be leveraged by a number of semi-supervised techniques \cite{zhu2023transductive, bendou2022easy}. In a transductive setting, the typical goal is to infer the classes of the query set all at once. However, in this context, we consider the performance on the separate test set mentioned earlier to account for better task generalization and allow for a better comparison with the inductive setting.
We have chosen to explore these two settings as they can present varying degrees of difficulty for locating the object of interest in relation to annotating an additional example. In the transductive setting, annotating a sample can be as simple as moving it from the query set to the support set, whereas in the inductive setting, labeling an additional sample requires further data acquisition which can be quite expensive in certain applications.

\section{Leveraging the location of the object of interest in the image}
\label{LevLoc}

Our goal is to differentiate between different types of physical objects (e.g., cat versus dog). However, models are usually trained on images containing multiple objects at once, leading to ambiguous representations and making generalization challenging. The significance of this correlation is likely to be discounted given a large dataset. In a one-shot setting, however, a spurious object may be considered with equal importance as the object of interest, thereby rendering additional information about the position of the object of interest within the image useful to disambiguate.

\subsection{Obtaining the location information}

We consider three different methods of acquiring the object's location with varying degrees of human involvement. Each method has its advantages and drawbacks, depending on the problem settings. 

\subsubsection{Fully Manual Human Annotation}

A direct approach to acquiring the object's location information is soliciting input from a human annotator. This approach involves querying the annotator for the most compact bounding box that entirely encompasses the object of interest. While this methodology is manual, it excels in delivering high-caliber bounding boxes, ensuring the comprehensive coverage of the object. While requiring such annotations in large classification datasets can be costly, we advocate that in many few-shot settings, obtaining bounding boxes of objects of interest might be less demanding. Moreover, in the inductive setting, this additional cost is justifiable as it reduces that of acquiring new data for the considered task.

\subsubsection{Using Segment Anything Model}

This second method for obtaining location information relies on sophisticated segmentation models. Specifically, leveraging the Segment Anything Model (SAM)~\cite{kirillov2023segment} enables the generation of segmentation for all objects within a given image. Integrating this model into our semi-manual approach involves soliciting a human annotator to pinpoint the object of interest within the image. Subsequently, we utilize SAM to generate a mask from this point. SAM can be prompted with a point in the image through its pixel coordinates to produce a binary mask of an object containing that point. The mask is then converted into a bounding box.
This method minimizes human involvement, requiring the annotator to simply identify a point rather than delineate a bounding box. However, it leads to additional computational costs primarily associated with SAM's image encoder and risks producing lower quality bounding boxes if an erroneous mask is generated.

\subsubsection{Automatic Salient Object Detection}

In this fully automatic mode, we rely on unsupervised object segmentation model MOVE~\cite{bielski2022move} to generate foreground object masks. MOVE leverages the fact that shifting foreground objects relative to their initial positions results in new, realistic images. It uses a combination of image inpainting and adversarial training to teach a model how to generate accurate foreground object masks. We utilize this model, trained in a fully unsupervised manner, to generate masks for the object of interest. While this method eliminates human involvement altogether, it also leads to additional computational costs associated with multiple image encoders. Furthermore, it has the major drawback of being unable to differentiate between the relevant object of interest and other objects that happen to be in the foreground of the image, unlike the other two methods.

\subsection{Methodology}

Our approach involves training a model on augmented versions of the images, with a specific focus on the object of interest. We leverage a large pre-trained model and adapt it to our task.
Our methodology revolves around few-shot runs, each comprising \(w\) classes, \(s\) annotated training samples, and \(t\) test samples. In the transductive scenario, we also have access to \(q\) unlabeled samples. For simplicity, we consider class-balanced pools of labeled, unlabeled, and test samples.

\subsubsection{Data Augmentation With Crops}
\label{dataaugcrop}
Assume we have access to bounding boxes for the labeled pool. The initial phase involves augmenting the support set using the bounding box information. For the ground truth bounding boxes, derived from fully manual human annotation, each image is augmented once with a crop around the object of interest plus a context window of 60 pixels (30 pixels from each side of the bounding box). In the case of fully automatic or SAM-generated bounding boxes, each image from the support set is augmented three times: first, by resizing the crop to encompass 20\% of the remaining context, then with 50\% of the remaining context, and finally to include 80\% of the context. The rationale behind these choices is elaborated further in Section~\ref{seccropusage}.
The resulting augmented set is then employed for subsequent training steps explained thereafter.

\begin{figure*}[ht!]

\pgfplotstableread[col sep = comma, header=true] {crop_res3.csv}\cropres
\begin{center}
\scalebox{0.8}{
\begin{tikzpicture}

	\begin{groupplot}[group style={
	  				group size=3 by 2,vertical sep=2cm},
			  height=0.25\textheight,width=0.45\textwidth]
		\nextgroupplot[title=ImageNet Inductive,legend style={nodes={scale=0.6, transform shape}},         legend pos=south east,ylabel= accuracy,xtick={5,10,15,20,25}]
			\addplot [blue,thick] table [x=z, y=a] \cropres;
                \addlegendentry{No Cropping}
                \addplot [green,thick] table [x=z, y=b] \cropres;
                \addlegendentry{SAM crops}
                \addplot [red,thick] table [x=z, y=c] \cropres;
                \addlegendentry{Ground truth crops}
                \addplot [gray,thick] table [x=z, y=s] \cropres;
                \addlegendentry{Salient object crops}
		\nextgroupplot[title=CUB Inductive,legend style={nodes={scale=0.6, transform shape}},         legend pos=south east,xtick={5,10,15,20,25}]
 	   		\addplot [blue,thick] table [x=z, y=m] \cropres;
                \addlegendentry{No Cropping}
                \addplot [green,thick] table [x=z, y=n] \cropres;
                \addlegendentry{SAM crops}
                \addplot [red,thick] table [x=z, y=o] \cropres;
                \addlegendentry{Ground truth crops}
                \addplot [gray,thick] table [x=z, y=w] \cropres;
                \addlegendentry{Salient object crops}
		\nextgroupplot[title=Pascal VOC Inductive,legend style={nodes={scale=0.6, transform shape}},         legend pos=south east,xtick={5,10,15,20,25}]
 	   		\addplot [blue,thick] table [x=z, y=g] \cropres;
                \addlegendentry{No Cropping}
                \addplot [green,thick] table [x=z, y=h] \cropres;
                \addlegendentry{SAM crops}
                \addplot [red,thick] table [x=z, y=i] \cropres;
                \addlegendentry{Ground truth crops}
                \addplot [gray,thick] table [x=z, y=u] \cropres;
                \addlegendentry{Salient object crops}                
		\nextgroupplot[title=ImageNet Transductive,legend style={nodes={scale=0.6, transform shape}},         legend pos=south east,ylabel= accuracy,xtick={5,10,15,20,25}]
  	  		\addplot [blue,thick] table [x=z, y=d] \cropres;
                \addlegendentry{No Cropping}
                \addplot [green,thick] table [x=z, y=e] \cropres;
                \addlegendentry{SAM crops}
                \addplot [gray,thick] table [x=z, y=t] \cropres;
                \addlegendentry{Salient object crops}                
                \addplot [red,thick] table [x=z, y=f] \cropres;
                \addlegendentry{Ground truth crops}
            \nextgroupplot[title=CUB Transductive,legend style={nodes={scale=0.6, transform shape}},         legend pos=south east,xlabel=Number of labeled samples,xtick={5,10,15,20,25}]
                \addplot [blue,thick] table [x=z, y=p] \cropres;
                \addlegendentry{No Cropping}
                \addplot [green,thick] table [x=z, y=q] \cropres;
                \addlegendentry{SAM crops}
                \addplot [gray,thick] table [x=z, y=x] \cropres;
                \addlegendentry{Salient object crops}
                \addplot [red,thick] table [x=z, y=r] \cropres;
                \addlegendentry{Ground truth crops}    
            \nextgroupplot[title=Pascal VOC Transductive,legend style={nodes={scale=0.6, transform shape}},         legend pos=south east,xtick={5,10,15,20,25}]
                    \addplot [blue,thick] table [x=z, y=j] \cropres;
                \addlegendentry{No Cropping}
                \addplot [green,thick] table [x=z, y=k] \cropres;
                \addlegendentry{SAM crops}
                \addplot [gray,thick] table [x=z, y=v] \cropres;
                \addlegendentry{Salient object crops}
                \addplot [red,thick] table [x=z, y=l] \cropres;
                \addlegendentry{Ground truth crops}   
	\end{groupplot}

\end{tikzpicture}
}
        \end{center}
\caption{Comparison of classification performance using our methodology between the augmentation with crops scenarios (Salient Object Detection, SAM-generated crops and ground truth) and the baseline scenario without any local information about the object of interest. Results are reported for all three considered datasets in both inductive and transductive setting. We compute results for a variable number of labeled samples (support set). These results were computed over 100 runs.}
\label{figcropres}
\end{figure*}
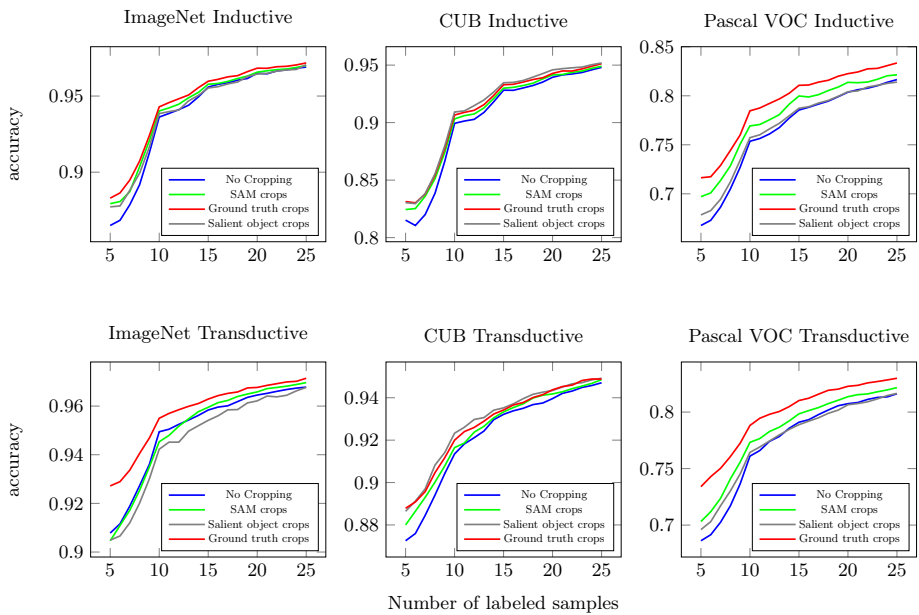

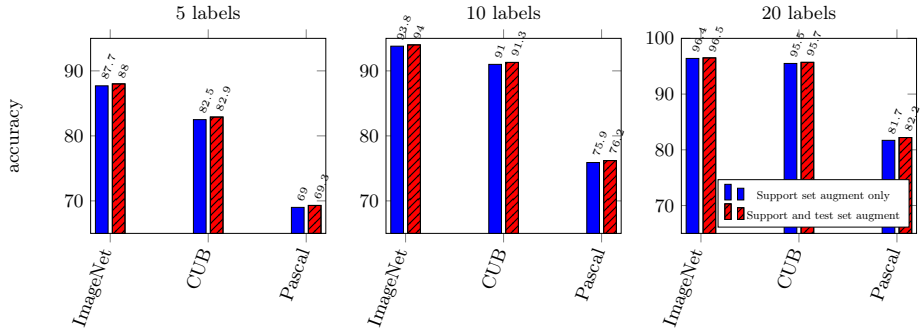
\begin{figure*}[t]
\scalebox{0.8}{
\begin{tikzpicture}

	\begin{groupplot}[group style={
	  				group size=3 by 1,vertical sep=1cm},
			  height=0.25\textheight,width=0.45\textwidth]

		\nextgroupplot[
                ybar,
                bar width=6pt,
                symbolic x coords={ImageNet,CUB, Pascal}, 
                x tick label style={anchor=east,rotate=70},
                nodes near coords,
                nodes near coords style={font=\tiny,anchor=west,rotate=70},
                ylabel=accuracy,
                ymin =65,
                ymax=95,
                title=5 labels, 
                xtick=data,
                legend style={nodes={scale=0.6, transform shape}},         legend pos=south east,ylabel= accuracy]
                    ]
                 \addplot[fill=blue] coordinates {(ImageNet, 87.7) (CUB, 82.5) (Pascal, 69.0) };
                \addplot[fill=red,postaction={pattern=north east lines}] coordinates {(ImageNet, 88.0) (CUB, 82.9) (Pascal, 69.3) };

  		\nextgroupplot[
                    ybar,
                    bar width=6pt,
                    symbolic x coords={ImageNet,CUB, Pascal}, 
                    x tick label style={anchor=east,rotate=70}, 
                    nodes near coords,
                    nodes near coords style={font=\tiny,anchor=west,rotate=70},
                ymin =65,
                ymax=95,
                    title=10 labels, 
                    xtick=data, 
                    legend style={nodes={scale=0.6, transform shape}},         legend pos=south east]
                        ]
                 \addplot[fill=blue] coordinates {(ImageNet, 93.8) (CUB, 91.0) (Pascal, 75.9) };
                  \addplot[fill=red,postaction={pattern=north east lines}] coordinates {(ImageNet, 94) (CUB, 91.3) (Pascal, 76.2) };

  		\nextgroupplot[
                    ybar,
                    bar width=6pt,
                    symbolic x coords={ImageNet,CUB, Pascal}, 
                    x tick label style={anchor=east,rotate=70},  
                                    ymin =65,
                ymax=100,
                    title=20 labels, 
                    xtick=data, 
                    nodes near coords,
                    nodes near coords style={font=\tiny,anchor=west,rotate=70},
                    legend style={nodes={scale=0.6, transform shape}},         legend pos=south east]
                        ]
                 \addplot[fill=blue] coordinates {(ImageNet, 96.4) (CUB, 95.5) (Pascal, 81.7) };
                  \addplot[fill=red,postaction={pattern=north east lines}] coordinates {(ImageNet, 96.5) (CUB, 95.7) (Pascal, 82.2) };

              \legend{Support set augment only,Support and test set augment}
	\end{groupplot}
\end{tikzpicture}
}
\caption{This figure illustrates the outcomes of augmenting the test set at inference time through salient object detection. The reported results pertain to all three datasets within 5-label, 10-label, and 20-label scenarios. A comparison is made between prediction results on an unaltered test set and those on an augmented test set using MOVE's automatic segmentation model. In both cases, training is performed with the MOVE-augmented support set. Results are based on 1000 runs.}
\label{saltesthist}
\end{figure*}

\begin{figure*}[ht!]

\pgfplotstableread[col sep = comma, header=true] {std_crop.csv}\cropstd
\scalebox{0.8}{
\begin{tikzpicture}

	\begin{groupplot}[group style={
	  				group size=3 by 1,horizontal sep=2cm},
			  height=0.1\textheight,width=0.20\textwidth]
		\nextgroupplot[title=ImageNet,legend style={nodes={scale=0.5, transform shape}},         legend pos=south east,ylabel= \textcolor{blue}{variance},axis y line=left,ytick pos=left,scale only axis,ticklabel style={font=\small}]
                
			\addplot [blue,thick] table [x=a, y=c] \cropstd;
               
            			\addplot [red,thick,draw=none] table [x=a, y=c] \cropstd;

		\nextgroupplot[title=CUB,legend style={nodes={scale=0.5, transform shape}},         legend pos=south east,xlabel = context percentage,axis y line=left,ytick pos=left,scale only axis,ticklabel style={font=\small}]
                
			\addplot [blue,thick] table [x=a, y=d] \cropstd;
               
            			\addplot [red,thick,draw=none] table [x=a, y=d] \cropstd;
 ;

		\nextgroupplot[title=Pascal ,legend style={nodes={scale=0.5, transform shape}},         legend pos=south east,axis y line=left,ytick pos=left,scale only axis,ticklabel style={font=\small}]
                
			\addplot [blue,thick] table [x=a, y=b] \cropstd;
               
            			\addplot [red,thick,draw=none] table [x=a, y=b] \cropstd;

	\end{groupplot}
    \begin{groupplot}[group style={
	  				group size=3 by 1,horizontal sep=2cm},
			  height=0.1\textheight,width=0.20\textwidth]
		\nextgroupplot[title=ImageNet,legend style={nodes={scale=0.3, transform shape}},         legend pos=south east,axis y line=right,ytick pos=right,axis x line = none,scale only axis,ticklabel style={font=\small}]
   
                \addplot [red,thick,dashed] table [x=a, y=e] \cropstd;

		\nextgroupplot[title=CUB,legend style={nodes={scale=0.6, transform shape}},         legend pos=south east,axis y line=right,ytick pos=right,axis x line = none,scale only axis,ticklabel style={font=\small}]

                    \addplot [red,thick,dashed] table [x=a, y=f] \cropstd;

		\nextgroupplot[title=Pascal ,legend style={nodes={scale=0.6, transform shape}},         legend pos=south east,ylabel style={align=centertext ,width=2cm},ylabel=\textcolor{red}{distance\newline (dashed)},axis y line=right,ytick pos=right,axis x line = none,scale only axis,ticklabel style={font=\small}]

                    \addplot [red,thick,dashed] table [x=a, y=g] \cropstd;

	\end{groupplot}
\end{tikzpicture}
}
\caption{Illustration of the impact of object-centric cropping on class variance. Our analysis encompasses the average across 100 classes for ImageNet, all 20 classes for Pascal VOC, and the 60 training classes of CUB. Across all datasets, we examine a random distribution of 100 samples for each class. We present the average class variance of the latent representations and average distance to the original uncropped class means for different percentages of context, where 0 corresponds to the minimal crop (i.e., the most compact bounding box), 1 represents the entire image, and intermediate values indicate linear interpolation between the two extremes.}
\label{figcroplat}
\end{figure*}
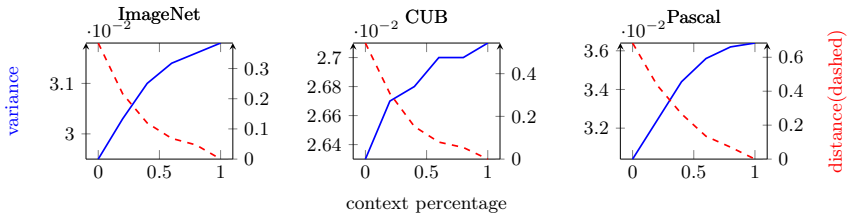

\begin{figure*}[ht!]
\centering

\begin{tikzpicture}
\pgfplotstableread[col sep = comma, header=true] {scatter_2.csv}\scattertwo
\pgfplotstableread[col sep = comma, header=true] {scatter_2c.csv}\scattertwoc
\pgfplotstableread[col sep = comma, header=true] {scatter_3.csv}\scatterthree
\pgfplotstableread[col sep = comma, header=true] {scatter_3c.csv}\scatterthreec

\begin{groupplot}[group style={
	  				group size=4 by 1,horizontal sep=0.5cm},
			  height=0.20\textheight,width=0.3\textwidth]
		\nextgroupplot[ymajorticks=false,xmajorticks=false,title=uncropped,legend style={nodes={scale=0.3, transform shape}},         legend pos=south east,ylabel= PCA dim 1,xlabel = PCA dim 2,ymin=-1.5,ymax=1,xmin=-1,xmax=1.5]
        
        \addplot [only marks, scatter, scatter src=explicit
        ,        scatter/classes={%
                0={mark=*,draw=blue,fill=blue,fill opacity=0.1,draw opacity=0.2},
                1={mark=*,fill opacity=0.1,draw opacity=0.2,draw=red,fill=red},
                2={mark=*,fill opacity=0.1,draw opacity =0.2,draw=green,fill=green}}, mark options={scale=1}] table [x=x, y=y, meta=c,mark=c] 
        \scattertwoc;

  \addplot[smooth,mark=diamond*,blue,mark options={scale=2}] 
  coordinates{(0.09,-0.61)};
\addplot[smooth,mark=diamond*,red,mark options={scale=2}] 
  coordinates{(0.91,-0.13)};

\node[] at (80,230) {\small 2 classes};

		\nextgroupplot[ymajorticks=false,xmajorticks=false,title=cropped,legend style={nodes={scale=0.3, transform shape}},     legend pos=south east,ymin=-1.5,ymax=1,xmin=-1,xmax=1.5]
        \addplot [only marks, fill opacity=0.1,draw opacity=0.2,scatter, scatter src=explicit, mark options={scale=1}] table [x=x, y=y, meta=c,mark=c] \scattertwo;

\addplot[fill opacity=0.5,smooth,mark=diamond*,blue,mark options={scale=2}] 
  coordinates{(-0.1,-0.37)};
\addplot[fill opacity=0.5,smooth,mark=diamond*,red,mark options={scale=2}] 
  coordinates{(0.75,-0.02)};

\addplot[fill opacity=0.1,smooth,mark=diamond*,blue,mark options={scale=2}] 
  coordinates{(0.09,-0.61)};
\addplot[fill opacity=0.1,smooth,mark=diamond*,red,mark options={scale=2}] 
  coordinates{(0.91,-0.13)};

\node[] (source) at (axis cs:0.09,-0.61){};
\node (destination) at (axis cs:-0.1,-0.37){};
\draw[->](source.center)--(destination.center);

\node[] (source1) at (axis cs:0.91,-0.13){};
\node (destination1) at (axis cs:0.75,-0.02){};
\draw[->](source1.center)--(destination1.center);

	\nextgroupplot[ymajorticks=false,xmajorticks=false,xshift = 1cm,title=uncropped ,legend style={nodes={scale=0.3, transform shape}},         legend pos=south east,ymin=-1.5,ymax=1.5,xmin=-1.5,xmax=1.5]
        \addplot [only marks,  fill opacity=0.1,draw opacity=0.2,scatter, scatter src=explicit,scatter/classes={%
                0={mark=*,draw=blue,fill=blue,fill opacity=0.1,draw opacity=0.2},
                1={mark=*,fill opacity=0.1,draw opacity=0.2,draw=red,fill=red},
                2={mark=*,fill opacity=0.1,draw opacity =0.2,draw=green,fill=green}}, mark options={scale=1}] table [x=x, y=y, meta=c,mark=c] \scatterthree;

        \addplot[smooth,mark=diamond*,blue,mark options={scale=2}] 
          coordinates{(0.91,-0.13)};
        \addplot[smooth,mark=diamond*,red,mark options={scale=2}] 
          coordinates{(-0.72,0.16)};
        \addplot[smooth,mark=diamond*,green,mark options={scale=2}] 
          coordinates{(0.1,-0.47)};

        \node[] at (100,270) {\small 3 classes};

        \nextgroupplot[ymajorticks=false,xmajorticks=false,xshift = 0cm,title=cropped ,legend style={nodes={scale=0.3, transform shape}},         legend pos=south east,ymin=-1.5,ymax=1.5,xmin=-1.5,xmax=1.5]
        
        \addplot [only marks,  fill opacity=0.1,draw opacity=0.2,scatter, scatter src=explicit,scatter/classes={%
                0={mark=*,draw=blue,fill=blue,fill opacity=0.1,draw opacity=0.2},
                1={mark=*,fill opacity=0.1,draw opacity=0.2,draw=red,fill=red},
                2={mark=*,fill opacity=0.1,draw opacity =0.2,draw=green,fill=green}}, mark options={scale=1}] table [x=x, y=y, meta=c,mark=c] \scatterthreec;

        \addplot[fill opacity=0.5,smooth,mark=diamond*,blue,mark options={scale=2}] 
          coordinates{(0.75,-0.02)};
        \addplot[fill opacity=0.5,smooth,mark=diamond*,red,mark options={scale=2}] 
          coordinates{(-0.6,0.04)};
        \addplot[fill opacity=0.5,smooth,mark=diamond*,green,mark options={scale=2}] 
          coordinates{(0,-0.33)};

        \addplot[fill opacity=0.1,smooth,mark=diamond*,blue,mark options={scale=2}] 
          coordinates{(0.91,-0.13)};
        \addplot[fill opacity=0.1,smooth,mark=diamond*,red,mark options={scale=2}] 
          coordinates{(-0.72,0.16)};
        \addplot[fill opacity=0.1,smooth,mark=diamond*,green,mark options={scale=2}] 
          coordinates{(0.1,-0.47)};

        \node[] (source) at (axis cs:0.91,-0.13){};
        \node (destination) at (axis cs:0.75,-0.02){};
        \draw[->](source.center)--(destination.center);
        
        \node[] (source) at (axis cs:-0.72,0.16){};
        \node (destination) at (axis cs:-0.6,0.04){};
        \draw[->](source.center)--(destination.center);
        
        \node[] (source) at (axis cs:0.1,-0.47){};
        \node (destination) at (axis cs:-0,-0.33){};
        \draw[->](source.center)--(destination.center);
	
    \end{groupplot}
 
 \end{tikzpicture}
 \caption{This figure offers a two-dimensional visualization of the feature space, showcasing class distributions for Pascal VOC. We display 20 random samples (circles) and their centroids (diamond) from two and three random Pascal VOC classes, contrasting the distribution of the latent representation of the uncropped images with the latent representation of the image crops. In the ``cropped'' instance we display again the uncropped centroids as lower opacity diamonds to better showcase the shift.}
 \label{scattercroplat}
\end{figure*}
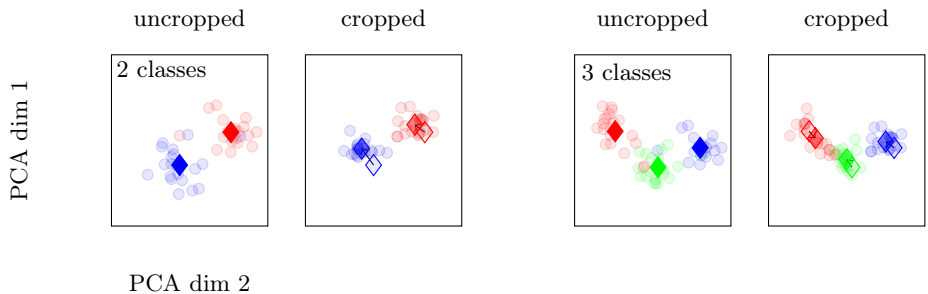

\label{saltest}

\begin{figure*}[t]

\begin{tikzpicture}

	\begin{groupplot}[group style={
	  				group size=3 by 1,vertical sep=2cm},
			  height=0.20\textheight,width=0.35\textwidth]
     
		\nextgroupplot[
                ybar,
                bar width=6pt,
                symbolic x coords={replace,minimal,+60,+150,20\%,50\%,multiple}, 
                x tick label style={anchor=east,rotate=70},
                ylabel=accuracy,
                ymin =50,
                ymax=100,
                title=ImageNet, 
                xtick=data,
                legend style={nodes={scale=0.6, transform shape}},         legend pos=south east,ylabel= accuracy]
                    ]

              \addplot[fill=blue] coordinates {(replace, 81.1) (minimal, 86.8) (+60, 88.6) (+150, 87.3) (20\%, 88.1) (50\%, 88.1) (multiple, 89.4)};
              
              \addplot[fill=red,postaction={pattern=north east lines}] coordinates {(replace, 89.1) (minimal, 91.5) (+60, 93.2) (+150, 93.5) (20\%, 93.5) (50\%, 93.6) (multiple, 94.3)};

  		\nextgroupplot[
                    ybar,
                    bar width=6pt,
                    symbolic x coords={replace,minimal,+60,+150,20\%,50\%,multiple}, 
                    x tick label style={anchor=east,rotate=70},  
                    ymin =40,
                    ymax=90,
                    title=CUB, 
                    xtick=data, 
                    legend style={nodes={scale=0.6, transform shape}},         legend pos=south east]
                        ]
                 \addplot[fill=blue] coordinates {(replace, 49.2) (minimal, 79.3) (+60, 82.7) (+150, 81.9) (20\%, 81.9) (50\%, 82.4) (multiple, 83.7)};
                  \addplot[fill=red,postaction={pattern=north east lines}] coordinates {(replace, 52.7) (minimal, 82.3) (+60, 84.4) (+150, 82.2) (20\%, 83.3) (50\%, 82.6) (multiple, 82.5)};

  		\nextgroupplot[
                    ybar,
                    bar width=6pt,
                    symbolic x coords={replace,minimal,+60,+150,20\%,50\%,multiple}, 
                    x tick label style={anchor=east,rotate=70},
                    ymin =30,
                    ymax=80,
                    title=Pascal VOC,
                    xtick=data,
                    legend style={nodes={scale=0.6, transform shape}},         legend pos=south east]
                        ]
                 \addplot[fill=blue] coordinates {(replace, 60.5) (minimal, 68.6) (+60, 71.6) (+150, 70.3) (20\%, 71.5) (50\%, 69.9) (multiple, 72.0)};
                  \addplot[fill=red,postaction={pattern=north east lines}] coordinates {(replace, 66.0) (minimal, 72.4) (+60, 72.5) (+150, 69.9) (20\%, 71.4) (50\%, 69.5) (multiple, 72.3)};

              \legend{SAM crops,ground truth crops}
	\end{groupplot}
\end{tikzpicture}
\caption{Comparison of different methods for augmenting the support set using bounding boxes centered on the object of interest across three datasets. This analysis considers both SAM-generated and ground truth bounding boxes, and the reported averages are based on 100 runs for a 5-class, 5-support samples, and 100-test samples.}
The X-axis is to be interpreted as follows:
\small
\begin{itemize}[noitemsep,topsep=3pt]
  \item replace: only the crop with an additional 60 pixels of context is used for training, discarding the original image
  \item minimal: augmenting the original image with a crop around the bounding box.
  \item +X: augmentation with a crop around the bounding box with an additional number of X context pixels in both width and height.
  \item X\%: augmentation with a resize of the crop that encompasses X\% of the remaining context between the whole image and the minimally augmented crop.
  \item multiple: three augmentations are used in addition to the original image: 20\%, 50\%, and 80\%.
\end{itemize}

\label{histcropusage}
\end{figure*}
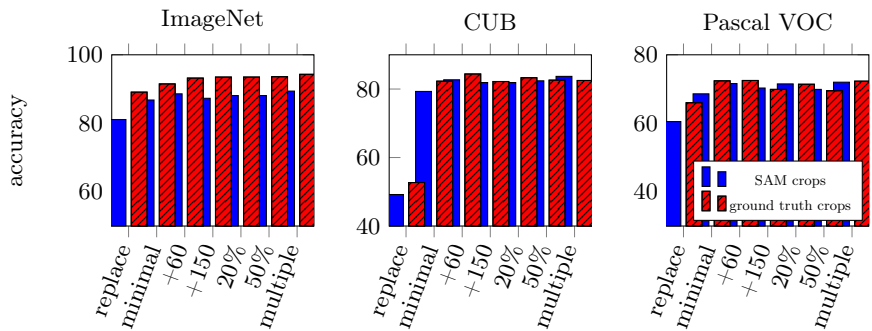
\subsubsection{Feature Extraction and Linear probing}

In this study, we chose to employ CLIP with a ResNet50 image encoder~\cite{radford2021learning} as a feature extractor and subsequently train a single linear layer atop it to tailor it to our classification task. This choice is motivated by recent works showing the ability of this combination to reach top-tier performance, while remaining almost hyperparameter free~\cite{lin2023multimodality,chen2020simple}.

\paragraph{In the inductive setting:} the linear layer is trained on the CLIP features from the support set and its augmented set.

\paragraph{In the transductive setting:} besides the extracted features from the support set and the augmentations, we also train on pseudolabels of the query set, which are generated as detailed in the subsequent section.

We employ a soft K-means algorithm for generating pseudolabels for the query set~\cite{bendou2022easy,abdali2023active}. The process commences by initializing \(w\) clusters with centroids as the class means computed from the support set and the augmentations. Subsequently, we iterate between assigning the elements from the query set to the clusters and computing the new clusters' centroids until convergence. The final cluster assignments are then utilized as the pseudolabels.

\label{ssec:subhead}

\section{Results}
\label{sec:majhead}
We conduct evaluations on both classification and object detection tasks. Unlike object detection datasets, classification benchmarks typically contain fewer objects per image, resulting in less distinct yet equally intriguing cases of ambiguity. This section outlines the datasets under consideration, the experimental setup, and showcases the added value of our methodology derived from harnessing local object information within the images. Additionally, we explore the impact of closely cropping around objects of interest on their representation in the feature space generated by the feature extractor. Lastly, we compare various approaches to incorporating object location information.
\subsection{Datasets}

We evaluate our method on an ImageNet~\cite{russakovsky2015imagenet} subset available with bounding box information in ImageNet Object Localization Challenge~\cite{imagenet-object-localization-challenge}, bird species dataset CUB following the split in ~\cite{chen2019closer}, and Pascal VOC which has 20 different classes of mainly common objects. While this dataset is typically considered for object detection, some works consider it for image classification tasks~\cite{shetty2016application}.

\subsection{Experimental setup}
\label{expset}

For consistency with many works in the literature on few-shot visual classification, we focus on few-shot tasks made of 
\(k=5\)
classes and a test set of
\(n_t=100\)
samples, along with a support set of
\(n_s=5\)
samples for the totality of our experiments. In the transductive scenario, we use a default setting where
\(n_s+n_q=50,\)
giving a total of 50 samples including labeled and unlabeled samples.

\subsection{Effects of cropping on task performance}

\label{seccropres}

For our inaugural experiment, our objective is to demonstrate the gains resulting from integrating the crops in our training. In Figure~\ref{figcropres}, we vary the number of support samples and contrast training solely on this support set with training on the augmented support set created using bounding boxes as we previously detailed in Section~\ref{dataaugcrop}.

Incorporating ground truth bounding boxes for augmentation leads to a clear improvement compared to baseline training. A notable 5\% increase is achieved for both inductive and transductive settings with 5 labeled samples in the case of Pascal VOC. This improvement is around 2\% for ImageNet and CUB. We attribute this difference in accuracy increase to Pascal VOC being an object detection dataset. It predominantly contains multiple object categories in most images as opposed to CUB and ImageNet. Consequently, the bounding box information in Pascal VOC offers more valuable insights for the model. Additionally, we observe a tendency for the gap between the baseline and ground truth curves to diminish as we increase the number of labeled samples, indicating that this disambiguation is more effective in lower shot settings.
Regarding the SAM-generated bounding boxes, we note that, for most cases, the curve lies between the two aforementioned curves. We attribute this to the quality of the mask generation. The salient object detection boxes yield very similar performances to those of SAM-generated or ground truth boxes in CUB and Imagenet, but result in slightly less, albeit still significant, performance boosts in Pascal VOC. We attribute this directly to the method's agnosticism to which object is of interest. In fact, when multiple objects are present in the foreground of an image, the generated mask can correspond to any of the foreground objects. Moreover, this experiment also emphasizes the associated return on our costs. For example, in the case of Pascal VOC in the inductive setting with only five labeled samples, enhancing these five samples results in a performance improvement similar to that achieved with eight total samples, a comparison we think to be valuable when deciding how to best leverage human annotation.

\subsection{Inference Time Disambiguation with Salient Object Detection}

In this section, we investigate the advantages of inferencing on specific crops of the test images as opposed to the entire image as a single instance. To conduct this experiment, we consider the inductive setting. We employ the training procedure based on the automatically generated masks. We then use the same automatic salient object detection to extract the prominent object from a test image at inference time. We then create various crops of this object at different sizes to generate multiple augmented images. Predictions are made for each of these augmented images, and the prediction with the highest confidence for its respective class is retained. We select the label corresponding to the most confident prediction among these augmented images, based on the linear head logits. Additionally, we establish a confidence threshold for the original test image. If the prediction confidence for the original image exceeds this threshold, the cropped images are disregarded, as high-confidence predictions likely reduce the need for further disambiguation.
As depicted in Figure~\ref{saltesthist}, this refined inference process leads to small improvements in classification accuracy across all examined support set sizes for all datasets. Although augmenting the test images can aid in disambiguation, in numerous instances, backgrounds from test images may arbitrarily align strongly with other incorrect classes, thus impeding performance. Another issue, similar to training, can arise when the automatically segmented object is not the object of interest.

\subsection{Impact of cropping on latent representations}
\label{seccroplat}

This second experiment seeks to elucidate the impact of cropping on the CLIP latent representation of images. In~Figure~\ref{figcroplat}, we compute the average class variance of the dataset's classes and the distance to the original (i.e. uncropped) class centroids for image crops with increasingly more context. Our observations reveal a consistent rise in variance as the contextual information increases contrasted with a decrease in the distance to the original class centroids. Notably, this trend is more pronounced for Pascal VOC. While the lower variance might help a model learn better, the shift in the class means creates a bias that hinders it. Hence the trade-off between minimizing cropping and retaining pertinent context.
Figure~\ref{scattercroplat} visually demonstrates this effect in a two-dimensional space. Here, we project CLIP features onto a two-dimensional space that retains the most variance for the uncropped dataset, achieved through Principal Component Analysis. Subsequently, we project random samples from Pascal VOC classes into this space. The resulting clusters corresponding to the classes appear more tightly knit in the case of the cropped samples, indicative of lower variance. Furthermore, we perceive the shift in the class centroids associated with the cropping.

\subsection{Augmentation with crops and context importance}
\label{seccropusage}

In this section, we explore different approaches to training the classifier using the ground truth and SAM-generated bounding boxes. we find that discarding the uncropped image can significantly hamper the model's performance. While the context can be spurious, it can also aid the model in adapting to the new domain. This effect is particularly evident in the ``replace'' mode for CUB in Figure~\ref{histcropusage}, where accuracy is nearly halved when discarding the original whole images.
Given this insight, the methodology was developed to preserve the original image and augment it with crops of various rescalings. We compare absolute resizing, which considers a fixed window in pixels around the bounding box, with relative resizing, where a percentage of the remaining context is factored into the augmentation generation. Additionally, we included a mode where we perform three augmentations with increasing relative context.
The results are quite close, as depicted in the Figure~\ref{histcropusage}. However, we observe that with ground truth boxes, the highest accuracy across datasets is achieved with a fixed context of 60 pixels, whereas for the SAM-generated bounding boxes, the ``multiple'' mode exhibits a slight advantage. This is likely due to flaws in SAM-generated bounding boxes, such as only covering a part of the object of interest, which is mitigated in the ``multiple'' mode.

\section{Conclusion}
\label{conc}
We have presented a novel approach to address the challenge of task ambiguity in the transductive and inductive scenarios of few-shot classification. Our methodology capitalizes on the local information of a target object within an image during the training phase, demonstrating strong performance on pertinent few-shot classification benchmarks. Additionally, we introduce a strategy that employs a large-scale segmentation model and a salient object detection model to reduce the human annotation cost associated with generating local object information. We believe this can bring value to many applications, especially industrial ones, where the data is ambiguous and costly to acquire. Looking ahead, we anticipate further exploration of efficient methodologies for identifying objects of interest within images.

\newpage


{\small
\bibliographystyle{ieee_fullname}
\bibliography{strings}
}

\vspace{12pt}

\color{red}

\end{document}